\RequirePackage{fix-cm}
\documentclass[twocolumn]{svjour3}          %
\smartqed  %
\usepackage{graphicx}
\usepackage{amsmath}
\usepackage{multirow}
\usepackage{pifont}
\usepackage{graphicx}
\usepackage{amssymb}
\usepackage{array}
\usepackage{multirow, nicefrac}
\usepackage{pifont}
\usepackage{cite}
\usepackage{bbm}
\usepackage{arydshln}
\usepackage{xcolor}
\usepackage{tablefootnote}
\usepackage{threeparttable}
\definecolor{green}{rgb}{1,0,0}

\def\etal{\emph{et al.}}

\def\etal{{\em et al.\/}\, }

\def\0{{\bf 0}}
\def\1{{\bf 1}}

\def\st{\mbox{s.t. }}

\usepackage{listings,xcolor} 
\usepackage{tcolorbox}

\newcommand{\lstfont}[1]{\color{#1}\ttfamily}

\usepackage{xspace}
\usepackage[colorlinks,linkcolor=red,anchorcolor=blue,citecolor=purple,CJKbookmarks=True]{hyperref}

\usepackage{marvosym}

\begin{document}

\title{Scaling Up 
Multi-domain Semantic Segmentation with Sentence Embeddings
}

\author{Wei Yin,
        Yifan Liu, %
        Chunhua Shen,
        Baichuan Sun,
        Anton van den Hengel
}
\authorrunning{\date, Accepted to Int. J. Computer Vision (IJCV)} %

\institute{
Wei Yin\at
              DJI \\
              \email{yvanwy@outlook.com}
          \and
          Chunhua Shen{\color{blue}{\Letter}} is the corresponding author.
          \at 
          Zhejiang University \\
          \email{chunhua@me.com} 
          \and 
          Baichuan Sun  \at  
          Amazon 
          \and
          Yifan Liu, Anton van den Hengel \at
          The University of Adelaide, Australia
}

\date{\today}

\maketitle

\maketitle

\begin{abstract} 
The state-of-the-art semantic segmentation methods have achieved impressive performance on predefined close-set individual datasets, but their generalization to zero-shot domains and unseen categories is limited. 
Labeling a large-scale dataset is challenging and expensive,
Training a robust semantic segmentation model on multi-domains has drawn much attention. 
However, inconsistent taxonomies hinder the naive merging of current publicly available annotations. To address this, we propose a simple solution to scale up the multi-domain semantic segmentation dataset with less human effort. We replace each class label with a sentence embedding, which is a vector-valued embedding of a sentence describing the class. This approach enables the merging of multiple datasets from different domains, each with varying class labels and semantics.
We merged publicly available noisy and weak annotations with the most finely annotated data, over 2 million images, which enables training a model that achieves performance equal to that of state-of-the-art supervised methods on $7$ benchmark datasets, despite not using any images therefrom. Instead of manually tuning a consistent label space, we utilized a vector-valued embedding of short paragraphs to describe the classes. 
By fine-tuning the model on standard semantic segmentation datasets, we also achieve a significant improvement over the state-of-the-art supervised segmentation on NYUD-V2~\cite{silberman2012indoor} and PASCAL-context~\cite{everingham2015pascal} at $60\%$ and $65\%$ mIoU, respectively.  
Our method can segment unseen labels based on the closeness of language embeddings, showing strong generalization to unseen image domains and labels. Additionally, it enables impressive performance improvements in some adaptation applications, such as depth estimation and instance segmentation. 

\keywords{Multi Domain, Sentence Embedding, Open World, Semantic Segmentation}

\end{abstract}

\section{Introduction}

Semantic segmentation is a fundamental task in computer vision with numerous applications in fields such as autonomous driving, agriculture robotics, and medicine. It serves as a precursor to several downstream applications, including scene understanding and a variety of image and video editing operations. Deep learning methods have achieved remarkable success in this task through supervised training on high-quality datasets with per-pixel labels for a limited set of classes~\cite{cordts2016cityscapes,lin2014microsoft}. However, these methods assume that all classes in testing are present during training, which is often not the case in real-world scenarios. Additionally, these models are constrained to the image domains represented in the training dataset, limiting their ability to generalize to new domains and labels. Although various approaches~\cite{xian2019semantic, hu2020uncertainty, baek2021exploiting, zhao2017open} have been proposed to address the open-set issues, they mainly conduct the experiment on a small dataset, which 
clearly limits its potential for real-world scenarios.

\begin{figure*}[t]
\setlength{\belowcaptionskip}{-0.5cm}
\centering  %
\includegraphics[width=1.5\columnwidth]{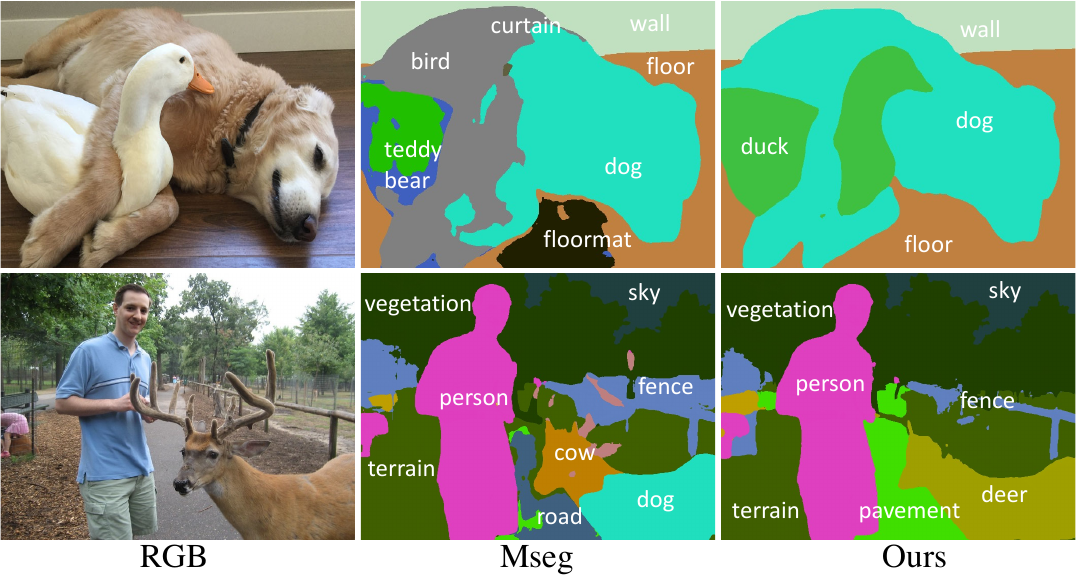}
\caption{Examples of segmenting zero-shot labels on web images. The labels `duck' (first row) and `deer' (second row) are not in the training data for Mseg~\cite{MSeg_2020_CVPR}, or the proposed method. Note that our proposed method is able to segment unseen labels successfully, however.   } 
\label{Fig: zero-shot labels cmp.}
\end{figure*}

Training on multiple datasets from diverse domains is a natural approach to enhance a model's robustness to variations in the statistics of data and improve generalization. However, naively merging datasets can lead to conflicting label taxonomies, as highlighted by Lambert~\emph{et al.}\cite{MSeg_2020_CVPR}. Lambert\emph{et al.} proposed a laborious solution to this problem, which involves manually creating a unified label set and re-labeling the merged dataset. This is particularly challenging when datasets have semantic classes that need to be split to achieve consistency. While their approach yields well-trained models that are robust to image-domain changes, it has limitations in open-set scenarios where predefined classes are insufficient for segmentation.

Rather than attempting to manually unify a diverse set of taxonomies, and the corresponding labeled instances, 
here 
we propose a method for 
\textit{automatically}
merging datasets by replacing the labels therein.
Specifically, for each category in each dataset, we identify a sentence describing the semantic meaning of the class.  These sentences can be manually generated, but in this work are %
retrieved 
from Wikipedia.  An embedding is then generated from each sentence using a language model.
In this work, we use the CLIP~\cite{radford2021learning} language model, but note that no image information is passed to the model, only 
the label sentences.
Using the resulting vector-valued sentence embeddings as labels allows multiple datasets to be merged without manual intervention or inspection. The primary benefit is that it enables training on more data, from a wider variety of domains, than any single dataset can support.

By merging 9 datasets, we gain access to about 
2 Million training images, which span multiple domains. These datasets have a variety of annotation styles, however, ranging from per-pixel labels to bounding boxes.  To exploit these varying annotations, we propose the heterogeneous losses that enable to leverage the noisy OpenImages \cite{OpenImagesSegmentation} and weakly annotated Objects365 \cite{shao2019objects365}. Our method not only significantly improves model performance and generalization ability to various domains, but also offers 
the advantage that \textit{the resulting model is able to generalize to unseen labels}.  This can be achieved simply by generating a new sentence describing the new class and calculating its vector-valued embedding. Applying this new vector label within the already trained model enables zero-shot segmentation of the corresponding class. Figure~\ref{Fig: zero-shot labels cmp.} shows some zero-shot semantic segmentation examples.

As mentioned, the sentence labels used to merge the datasets are retrieved from Wikipedia.
The corresponding increase in the volume of training data available is what drives the improvement in both zero-shot and fully supervised performance to the point where both are significantly beyond the current state-of-the-art methods.
Figure~\ref{Fig: similarity map.} shows the benefit of sentence-valued class labels even when the classes themselves do not overlap.  The block-diagonal nature of the similarity matrix for sentence label embeddings illustrates the fact that this approach better identifies semantic similarities between classes than using an embedding of a single-word class label.
Recent co-current work, Lseg~\cite{li2022languagedriven}, proposes to embed the CLIP~\cite{radford2021learning} model in their system to solve the zero-shot classes segmentation problem. Owing to the robust image feature from CLIP, they can achieve impressive performance on unseen labels. In contrast, we focus on not only the unseen categories segmentation problem but also zero-shot domain generalization. Therefore, we propose to leverage sentence embeddings to solve mix-data taxonomy, which can easily scale up the training data. We train the model on a larger dataset rather than improving the techniques used in Lseg. 
Our method is able to exploit diverse datasets in training, which generates state-of-the-art zero-shot and fully-supervised performance as a result. The resulting model boosts the performance of related applications such as depth estimation and instance segmentation. Our main contributions are thus as follows.
\vspace{-0.2 em}
\begin{itemize}

    \item We propose a method for easily merging multiple datasets together by replacing labels with vector-valued sentence embeddings, and use the method to construct a large semantic segmentation dataset %
    of 
    around 2 million images.
    \item The semantic segmentation model trained on this combined dataset achieves state-of-the-art performance on 4 zero-shot datasets. When applied in a  supervised setting (by fine-tuning), we %
    surpass the state-of-the-art methods by over 5\% on Pascal Context~\cite{everingham2015pascal} and NYUDv2~\cite{silberman2012indoor}. Furthermore, our methods show the significant advantage of segmenting  unseen labels.
    \item To solve the unbalanced annotations quality of mixed datasets, we propose a heterogeneous loss to %
    accommodate 
    the label noise from OpenImages~\cite{OpenImagesSegmentation} and leverage weakly annotated Objects365~\cite{shao2019objects365}.
    \item With the created pseudo mask labels from our model, the monocular depth estimation accuracy is improved consistently over 5 zero-shot testing datasets. Furthermore, we can surpass the fully-supervised instance segmentation method with only 25\% labeled images on the COCO dataset~\cite{lin2014microsoft}.
\end{itemize}

\section{Related Work}
\noindent\textbf{Semantic Segmentation.}
Semantic segmentation requires per-pixel semantic labeling, 
and as such represents one of the fundamental problems in computer vision.  
The first fully convolutional approach proposed was FCN \cite{long2015fully}, and various schemes have been developed to improve upon it. The constant increase in network capacity has driven ongoing improvements in performance on public semantic segmentation benchmarks, driven by developments such as ResNet and Transformer.
Recently, stronger transformer backbones such as Segformer \cite{xie2021segformer}, Swin~\cite{liu2021Swin} have shown promises
for semantic segmentation. 
Researchers have developed various segmentation models by, \textit{e.g.}, investigating larger receptive fields, and exploiting contextual information.

\begin{figure*}[!ht]
\setlength{\belowcaptionskip}{-0.4cm}
\centering  %
\includegraphics[width=0.95\textwidth]{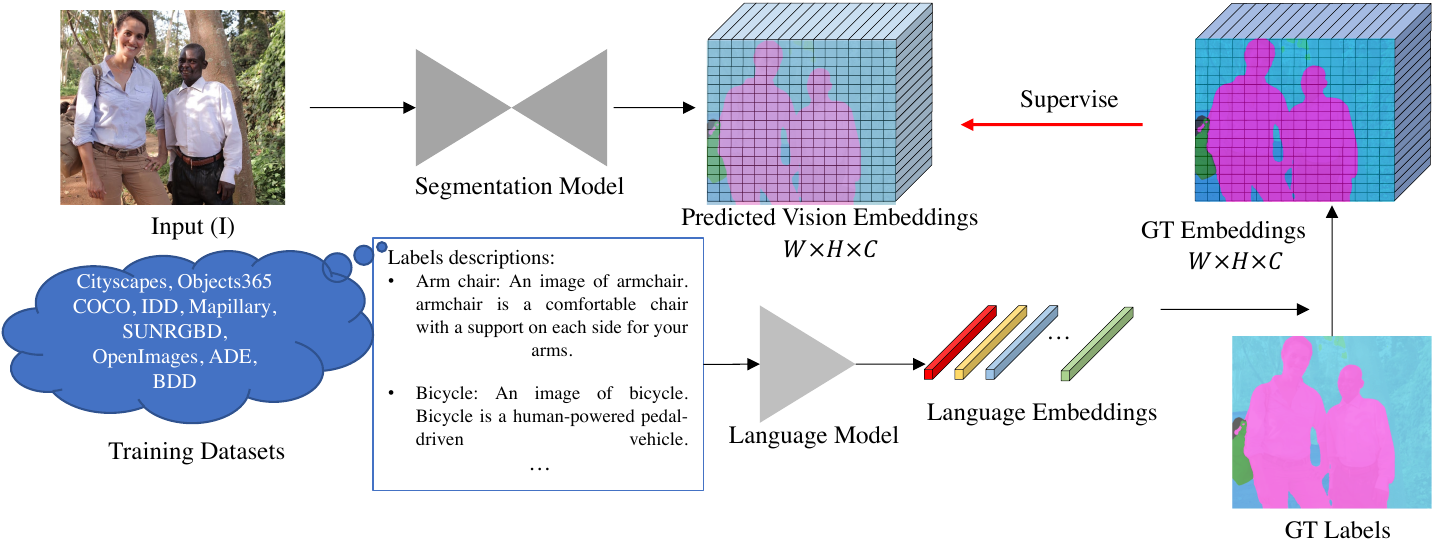}
\caption{\textbf{Our framework}. We merge multiple datasets together and encode all labels to embeddings %
using 
a language model. The semantic segmentation model is enforced to learn pixel-wise embeddings. During testing, the cosine similarity between the predicted embedding and the search embedding space %
is 
calculated. The %
output  
class is the most similar one in the label space.} 
\label{Fig: Framework}
\end{figure*}

\noindent\textbf{Zero-shot Semantic Segmentation.}
The term zero-shot has multiple meanings, including both applying a model to a dataset that it has not been trained on~\cite{MSeg_2020_CVPR}\cite{zendel2018wilddash}, and using a pre-trained model to identify novel classes~\cite{xian2019semantic}\cite{hu2020uncertainty}. The first category focuses on solving the zero-shot domain transferring problem. Most current semantic segmentation methods~\cite{cheng2021maskformer, cheng2021mask2former} train the model on a specific dataset, thus with limited generalization ability to diverse scenes. To solve this problem, Mseg proposes a  unified taxonomy to set up large-scale data for training, while wildash~\cite{zendel2018wilddash} proposes an evaluation benchmark.
Furthermore, how to segment the categories that have never been seen during training is a relatively new research topic~\cite{hu2020uncertainty, baek2021exploiting}. 
Existing zero-shot categories segmentation methods can be categorized into 2 streams, i.e. discriminative methods~\cite{zhao2017open, hu2020uncertainty, baek2021exploiting} and generative methods~\cite{bucher2019zero, gu2020context, ding2021decoupling}. 
Xian~\etal~\cite{xian2019semantic} proposed a discriminative zero-shot semantic segmentation method, which transfers each pixel's feature to a semantic word embedding space and obtains the class probability via the distance to the fixed semantic word embedding. 
Similarly, Zhao~\etal~\cite{zhao2017open} proposed a discriminative method to solve the zero-shot semantic segmentation problem using a label hierarchy. They thus incorporate hypernym/hyponym relations from WordNet~\cite{miller1995wordnet} to help with zero-shot parsing. Hu~\etal~\cite{hu2020uncertainty} argued that the noisy training data from seen classes
challenge the zero-shot label transfer. To address this issue, they proposed  an approach based on Bayesian uncertainty estimation.
Baek~\etal~\cite{baek2021exploiting} proposed to align the visual and semantic information through a learned joint embedding space and introduce boundary-aware regression and semantic consistency losses to learn discriminative features.
In contrast, the ZS3Net~\cite{bucher2019zero} is a typical generative method. They propose a generative model to generate per-pixel features by word embeddings and train it in a supervised manner. As the text embeddings are designed to describe the objects instead of pixels, ZS3Net's formulation is not robust. To create diverse and context-aware features from semantic word embeddings, Gu~\etal~\cite{gu2020context} proposed a contextual module to capture the pixel-wise contextual information in CaGNet.
A variety of zero-shot methods have followed~\cite{hu2020uncertainty}\cite{miller1995wordnet}\cite{bucher2019zero}. Existing methods are primarily either generative or discriminative.
In contrast to these pixel-wise zero-shot classification methods, Ding~\etal~\cite{ding2021decoupling} proposes decoupling the mask segmentation and multi-class classification problems. They first implement a class-agnostic segmentation and then combine it with CLIP~\cite{radford2021learning} to solve the category classification.

The approach that we propose is applicable to both versions of the zero-shot problem.
We demonstrate robust performance in both settings.  It is most notable, however, that our method achieves a level of accuracy on each of the $7$ major datasets, without training on them, that matches that of the fully-supervised approaches.

\noindent\textbf{Domain-agnostic Dense Prediction.}
There have been a range of methods that merge segmentation datasets to improve performance and generalization, including Ros~\etal~\cite{ros2016training} who %
combined  
six driving datasets, and Bevandic~\etal~\cite{bevandic2019simultaneous} who aggregated four datasets for joint segmentation and outlier detection on WildDash~\cite{zendel2018wilddash} (a benchmark designed for cross-domain robustness evaluation). As mentioned above,  Lambert~\etal~\cite{MSeg_2020_CVPR} proposed a method for creating a consistent taxonomy to unite datasets from multiple domains. They achieve strong generalization over multiple zero-shot datasets. 
Apart from semantic segmentation, the generalization to diverse scenes is also an issue for monocular depth estimation. Recent methods try to solve it by leveraging better learning objectives~\cite{xian2018monocular}\cite{chen2016single}\cite{yin2020diversedepth}\cite{metric3d}\cite{yin2022towards} for mixed datasets training~\cite{Ranftl2020}\cite{Wei2021CVPR}. They have achieved impressive performance in zero-shot domain transfer settings.

\noindent\textbf{Encoding Labels for Zero-shot Learning.}
Many zero-shot methods have generated semantic embeddings for class labels, and multiple methods for mapping between semantic embeddings and visual features have been devised. For example, Bucher~\etal~\cite{bucher2019zero} use \\ Word2Vec~\cite{mikolov2013distributed} to encode the labels, and is the only example that 
we are aware of in semantic segmentation. 
Recent examples that apply this approach for other purposes include Bujwid~\etal~\cite{bujwid2021large} and CLIP~\cite{radford2021learning}. A co-current work LSeg also uses language embeddings to supervise the class labels in semantic segmentation.

\section{Proposed Method}
\begin{figure*}[t]
\setlength{\belowcaptionskip}{-0.4cm}
\centering  %
\includegraphics[width=1.5\columnwidth]{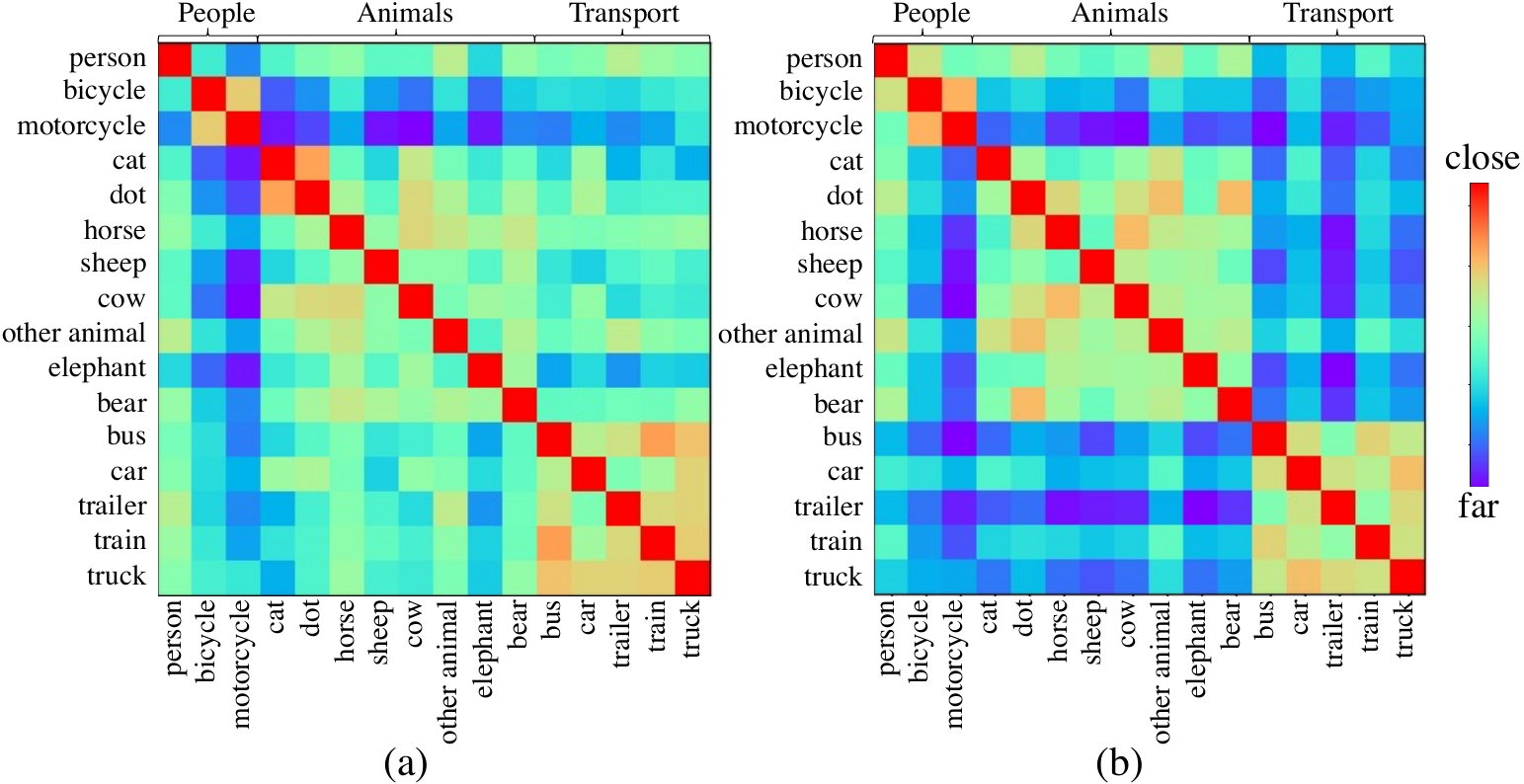}
\caption{A visualisation of the pairwise distances between label embeddings for (a) single word class labels and (b) sentences. The sentence embeddings better reflect the underlying semantic similarity between classes. For example, `Animals' classes are close to each other but far from those in `People' and `Transport'. Capturing and exploiting these semantic similarities in diverse datasets can improve both zero-shot, and fully supervised segmentation performance.} 
\label{Fig: similarity map.}
\end{figure*}

\def\cR{{\cal R}}

Figure~\ref{Fig: Framework} %
shows 
the overview of our method. Current semantic segmentation methods feed an RGB image $I\in {\cal R}^{W\times H\times 3}$ to a neural network and predict per-pixel classes $p \in {\cal R}^{W\times H\times N}$, where $N$ is the number of predetermined classes.

Recently, CLIP~\cite{radford2021learning} shows promising robustness by leveraging natural language supervision for image %
classification 
on a large-scale dataset. Inspired by this, instead of using one-hot vectors to represent the predefined labels, we employ language models to encode semantic labels into  embedding vectors.  In our experiments, we use the CLIP-ViT language model to encode all labels. The encoded language embeddings are $e \in {\cR}^{N\times C}$, where $C$ is the embedding dimension. The vision embeddings from the segmentation model are $\mathcal{V} = \phi(I) \in \cR^{W\times H\times C}$. 

\begin{figure*}[]
\centering  %
\includegraphics[width=0.65\textwidth]{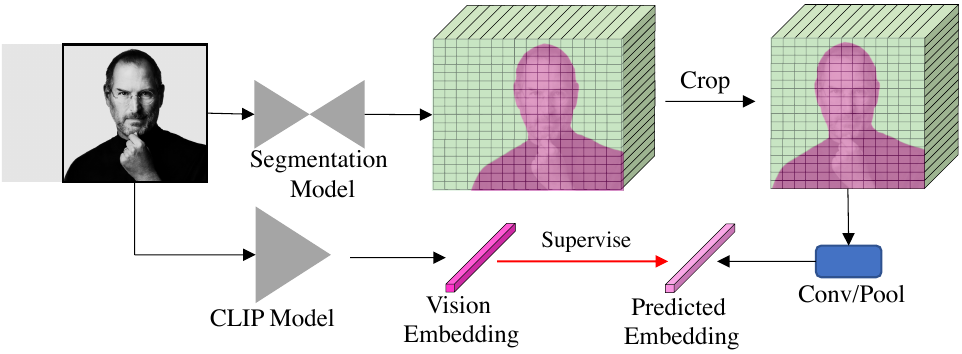}
\caption{Distillation pipeline for weakly-labeled data. We use the clip classification model to obtain the vision embedding for the bounding box regions.  It is applied to supervise the predicted bounding box embedding.} 
\label{Fig: distillation map.}
\end{figure*}

\noindent\textbf{Creating Language Embeddings.}
To train a robust model, we merged $9$ datasets together for training, containing $238$ labels in total. Lambert~\etal~\cite{MSeg_2020_CVPR} manually create a unified label list and encode each label in a one-hot vector. All labels are independent to each other. In contrast, we employ a soft semantic embedding to replace the original labels, which can retain the relations among labels. Such relations are important for segmenting zero-shot labels.

Ideally, the closeness between embeddings is positively correlated to the similarity between different labels. For example, `Pedestrian' should be close to `Rider' but far from `Animals' and other objects. Recently, several zero-shot semantic segmentation and classification methods~\cite{bucher2019zero}\cite{baek2021exploiting}\cite{hu2020uncertainty} explore pre-trained linguistic semantic features using class names (\textit{i.e.}, word2vec). The problem is that words are sensitive to linguistic issues and can give little discriminability of classes. Some names may partially overlap or not well reflect semantic similarity. For example, `counter' can be a counting device or a long flat-topped fitment. By contrast, textual descriptions are rich in context information. We thus collect  short descriptions from Wikipedia to represent each label. For example, `bus: An image of bus. A bus is a road vehicle designed to carry many passengers.' Figure~\ref{Fig: similarity map.} visualizes the cosine similarity matrix of embeddings, which are created from words or  sentences.  The  block-diagonal  nature demonstrates that our sentences embeddings better represent the semantic similarities. For example, classes in `Animals' are close to each other but far from those in `People'.

\noindent\textbf{Heterogeneous Constraints for Mixed Data Training.} 
To obtain a robust segmentation model, we merge 9 datasets for training, including 7 well-annotated datasets, a coarsely-annotated dataset (OpenImages~\cite{OpenImagesSegmentation}), and a weakly annotated dataset (Objects365~\cite{shao2019objects365}). Owing to the unbalanced quality, we propose to %
employ 
heterogeneous losses %
to train the model. 
For well annotated datasets, we enforce a pixel-wise loss on all samples. The loss function is %
as follows. 
\begin{align}
\begin{split}
    z_{i, j} = \frac{e_{j}\cdot \mathcal{V}_{i}}{\left \| e_{j} \right \|\left \|  \mathcal{V}_{i}\right \|}, i\in[0, M], j\in[0, N]
\end{split}\\
\begin{split}
    l_{i} = \frac{\exp(z_{i,j}/\tau )}{\sum_{j=0}^{N} \exp(z_{i,j}/\tau )}
\end{split}\\
\begin{split}
    L_{\rm HD}=-\frac{1}{M}
    \sum_{i=0}^{M}y_{i} \log(l_{i})
\end{split}
\end{align}
where $M$ is the number of %
valid pixels, $N$ total categories, $y_i$ is the ground-truth labels, and $\tau$ is the learnable temperature. $e$ is the created language embeddings of training labels, while $\mathcal{V}$ is the predicted vision embeddings.

\begin{figure*}[t]
\setlength{\belowcaptionskip}{-0.2cm}
\centering  %
\includegraphics[width=0.9\textwidth]{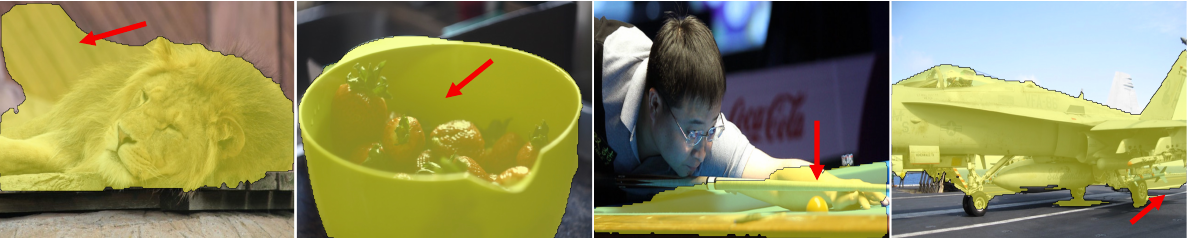}
\caption{Examples of noisy annotations from OpenImages (see red arrows regions). Left to right: masks for lion, bowl, billiard table, and airplane.} 
\label{Fig: noisy annotations.}
\vspace{-1 em}
\end{figure*}

As a coarsely annotated dataset contains many noises, we observe that directly enforcing the above pixel-wise loss leads to worse results. Figure~\ref{Fig: noisy annotations.} shows some noisy annotations from the OpenImages dataset. To alleviate the effect of such noises, we propose to enforce the loss on some high-confident samples and ignore the noisiest parts. The loss functions are 
as follows:
\begin{equation}
w_i= \begin{cases}
0,\quad & \log(l_i) > \mu \\
1,\quad &  \log(l_i) \leq \mu
\end{cases} 
\end{equation}

\begin{equation}
    L_{\rm LD} = -\frac{1}{M}\sum_{i=0}^{M}w_i\cdot  y_{i}  \log(l_{i}),
\end{equation}
where $\mu$ is an adaptive threshold. In our experiments, we rank all samples' losses of each image and set $\mu$ to the highest 30\% loss value. Each image has an adaptive threshold. With the proposed heterogeneous loss for OpenImages, we can obtain much better segments (see %
Figure~\ref{Fig: Effect of heterogeneous loss}).

Furthermore, we propose a distillation method to leverage the weakly annotated dataset, Ob\-je\-ct\-s\-365, wh\-i\-c\-h only contains bounding boxes for foreground objects. Specially, since the CLIP classification model has obtained robust knowledge of image contents, we distill such knowledge to our segmentation model on box-level annotations. 

We crop and resize the bounding box from the image $I_{r} = (crop(I, r))$ and feed it to the CLIP classification model to obtain the vision embedding for objects, $\mathcal{V}^{*}_r = \mathcal{G}(I_r)$. $I$ is the image and $r$ is the bounding box. To retrieve the predicted bounding box embedding, we crop the corresponding region $crop(\mathcal{V}, r)$ from the segmentation embeddings and enforce a $1 \times 1$ convolution and ROI pooling on that $\mathcal{V}_r = %
{\rm RoI} 
(crop(\mathcal{V}, r))$.  Lastly, we apply 
the $ \ell_1$
loss to minimize their distance. The pipeline is 
shown 
in Figure~\ref{Fig: distillation map.}.

\begin{small}
\begin{align}
\begin{split}
    \textbf{v}^{*}_r = \frac{\mathcal{V}^{*}_r}{\left \|\mathcal{V}^{*}_r\right \|}, \textbf{v}_r = \frac{\mathcal{V}_r}{\left \|\mathcal{V}_r\right \|}
\end{split}\\
\begin{split}
    L_{\rm WD} = \frac{1}{P}\sum_{r}^{P}\left \| \textbf{v}^{*}_r - \textbf{v}_r \right \|_{1}
\end{split}
\end{align}
\end{small}
where $P$ is the number of bounding boxes. The overall loss is as follows:
\begin{equation}
    L = L_{\rm HD} + L_{\rm LD} + L_{\rm WD}, 
\end{equation}
During inference, we calculate the cosine similarity between predicted vision embeddings and created language embeddings to obtain pixel-wise labels.

\section{Experiments}
\subsection{Datasets and Implementation Details}
In our experiments, we merged multiple datasets together for training and tested on 13 zero-shot datasets to evaluate the robustness of our method. 
 
\noindent\textbf{Training Data.} We firstly merged 7 high-quality annotated semantic segmentation datasets (HD), including ADE20K~\cite{zhou2017scene}, Mapillary~\cite{neuhold2017mapillary}, COCO Panoptic~\cite{lin2014microsoft}, India Driving Dataset (IDD)~\cite{varma2019idd}, BDD100K (BDD)~\cite{bdd100k}, Cityscapes~\cite{cordts2016cityscapes}, and SUNRGBD~\cite{song2015sun}. Apart from that, we sample some data from the instance segmentation dataset OpenImagesV6~\cite{OpenImagesSegmentation} (OI) and weakly annotated dataset Objects365~\cite{shao2019objects365} (OB) for training. Note that OpenImagesV6 only contains noisy foreground objects masks, while Objects365 provides objects bounding box annotations. These high-quality datasets (HD) have over 200K images. OpenImagesV6 has around 700K images, while Objects365 has over 1M images. 

\begin{table}[]
\centering
\resizebox{1.0\linewidth}{!}{%
\begin{tabular}{ll||ll}
\hline \hline
\multicolumn{2}{l||}{\begin{tabular}[c]{@{}c@{}}Testing data for\\ Semantic Segmentation\end{tabular}} & \multicolumn{2}{c}{\begin{tabular}[c]{@{}c@{}}Testing Data for\\ Downstream Applications\end{tabular}} \\ \hline
CamVid~\cite{brostow2008segmentation}                                            & Test set                                          & \multicolumn{2}{c}{Depth Estimation}                                                                   \\ \cline{3-4} 
Pascal VOC~\cite{everingham2015pascal}                                        & Validation set                                    & NYUv2                                          & Test set                                              \\
Pascal Context~\cite{mottaghi2014role}                                    & Validation set                                    & KITTI                                    & Test set                                              \\
KITTI~\cite{Geiger2013IJRR}                                             & Test set                                          & DIODE~\cite{diode_dataset}                                          & Test set                                              \\
Wilddash1~\cite{zendel2018wilddash}                                         & Validation set                                    & ScanNet~\cite{dai2017scannet}                                        & Validation set                                        \\
Wilddash2~\cite{zendel2018wilddash}                                         & Test set                                          & Sintel~\cite{Butler:ECCV:2012}                                        & Validation set                                        \\ \cline{3-4} 
Youtube VIS~\cite{yang2019video}                                       & Sampled data                                      & \multicolumn{2}{c}{Instance Segmentation}                                                              \\ \cline{3-4} 
NYUv2~\cite{silberman2012indoor}                                             & Test set                                          & COCO~\cite{lin2014microsoft}                                           & Validation set                                        \\ \hline\hline
\end{tabular}}
\caption{Datasets employed in our experiments for evaluation. \label{Tab: All test datsets}}
\end{table}

\begin{figure*}[!h]
\setlength{\belowcaptionskip}{-0.2cm}
\centering  %
\includegraphics[width=0.98\textwidth]{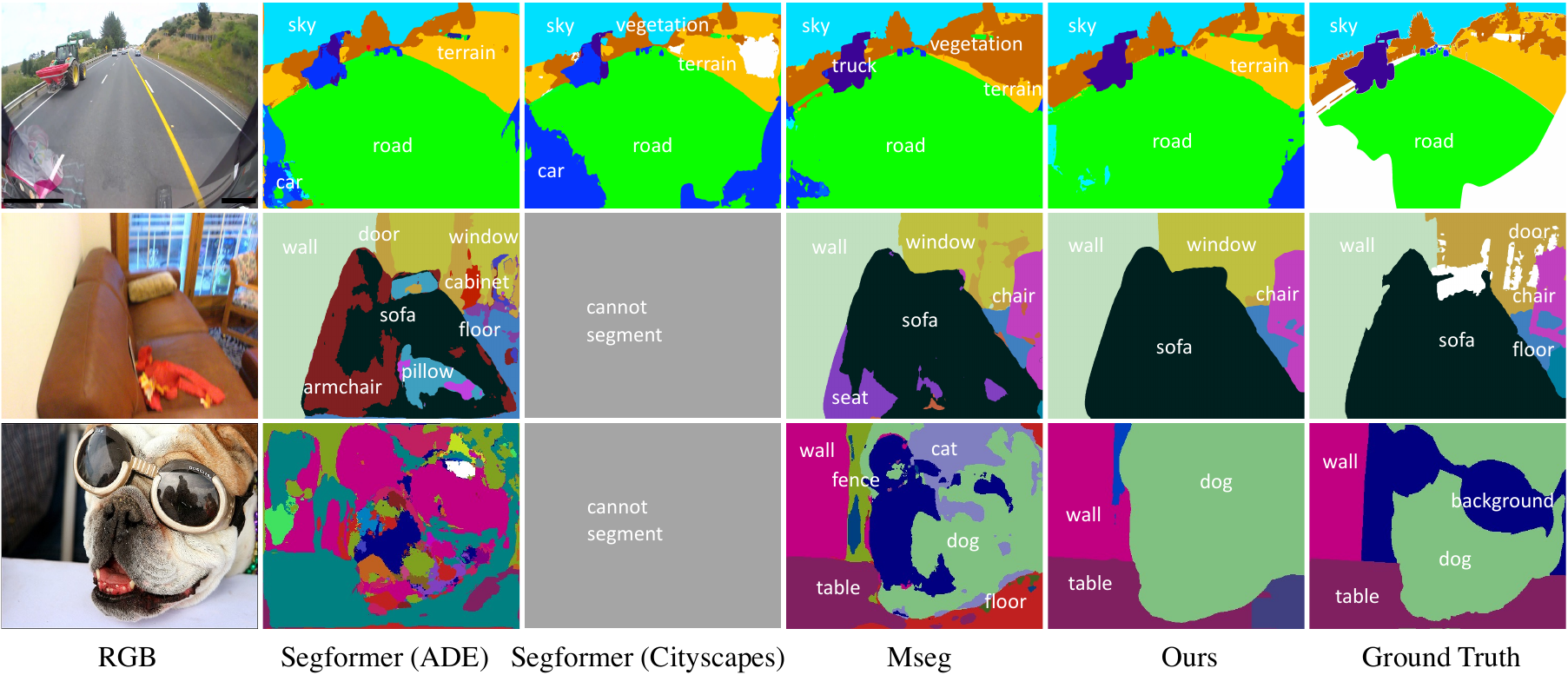}
\caption{Qualitative comparison on some zero-shot datasets (Pascal Context, ScanNet, and Wilddash1). Comparing with current methods, our method can retrieve better results.} 
\label{Fig: Cmp with SOTA}
\end{figure*}

\noindent\textbf{Testing Data for Semantic Segmentation.} To evaluate the robustness and effectiveness of our semantic segmentation method, we test on 8 zero-shot datasets, including CamVid~\cite{brostow2008segmentation}, 
KITTI, %
 Pascal VOC,
 Pascal Context,
 ScanNet~\cite{dai2017scannet}, WildDash1~\cite{zendel2018wilddash},  WildDash2~\cite{zendel2018wilddash}, YoutubeVIS~\cite{yang2019video}. Furthermore, we fine-tune our well-trained model on NYUv2~\cite{silberman2012indoor} and Pascal Context  to show our method can provide strong performance. 

\noindent\textbf{Testing Data for Downstream Applications.} We create pseudo-semantic labels on zero-shot datasets to boost downstream applications, including instance segmentation and monocular depth estimation. We create pseudo instance masks on Objects365 to help instance segmentation on COCO~\cite{lin2014microsoft}. For depth estimation, following LeReS~\cite{Wei2021CVPR}, we create pseudo semantic masks on 9 RGBD datasets, of which 4 datasets are employed for training and others are for zero-shot depth evaluation, including NYUDv2, KITTI, DIODE~\cite{diode_dataset}, ScanNet, and Sintel~\cite{Butler:ECCV:2012}.  More details are shown in Table~\ref{Tab: All test datsets}.

\begin{table*}[]
\centering
\resizebox{1\linewidth}{!}{%
\begin{tabular}{l|cccccc|c}
\hline\hline
Methods            & CamVid & KITTI & VOC  & Pascal Context & ScanNet & WildDash1 & H-Mean \\ \hline \hline
\multicolumn{8}{c}{State-of-the-art methods on corresponding benchmarks}    \\ \hline \hline
Zhu~\etal~\cite{zhu2019improving}  & 81.7   & -     & -    & -              & -       & -         & -    \\
Chroma UDA~\cite{erkent2020semantic}         & -      & 60.4  & -    & -              & -       & -         & -    \\
AutoDeepLab~\cite{liu2019auto}        & -      & -     & \textbf{82.0} & -              & -       & -         & -    \\
DeepLabV3+~\cite{chen2018encoder}         & -      & -     & -    & \textbf{54.5}           & -       & -         & -    \\
Valada~\etal~\cite{valada2020self}    & -      & -     & -    & -              & 52.9    & -         & -    \\ \hline \hline
\multicolumn{8}{c}{HRNet trained on a single dataset, zero-shot testing}    \\ \hline \hline
HRNet~\cite{SunXLW19} (COCO)       & 56.6   & 48.2  & 73.7 & 43.1           & 33.9    & 38.9      & 46.0 \\
HRNet (ADE)        & 53.5   & 44.3  & 34.6 & 24.0           & 43.8    & 37.0      & 37.1 \\
HRNet (Mappilary)  & 82.5   & 68.5  & 22.0 & 13.5           & 2.1     & 55.2      & 9.2  \\
HRNet (IDD)        & 70.5   & 50.7  & 14.5 & 6.3            & 1.6     & 40.6      & 6.5  \\
HRNet (BDD)        & 71.0   & 55.0  & 13.5 & 6.9            & 1.4     & 52.1      & 6.1  \\
HRNet (Cityscapes) & 65.3   & 58.1  & 12.1 & 6.5            & 1.7     & 30.1      & 6.7  \\
HRNet (SUN)        & 0.1    & 0.7   & 10.2 & 4.3            & 42.2    & 1.4       & 0.3  \\ \hline \hline
\multicolumn{8}{c}{Methods trained on mixed datasets, zero-shot testing}    \\ \hline \hline
MSeg~\cite{MSeg_2020_CVPR}               & 82.4   & 62.4  & 70.8 & 45.2           & 48.4    & 64.2      & 59.6 \\
Ours\_HRNet        & 83.4   & 66.9  & 74.9 & 48.6           & 52.0    & 64.0      & 62.6 \\
Ours\_Segformer    & \textbf{83.7}   & \textbf{68.9}  & \underline{81.1} & \underline{54.2}           & \textbf{55.3}    & \textbf{69.7}      & \textbf{66.9}     \\ \hline\hline
\end{tabular}}\\
\caption{Quantitative comparison to existing methods on 6 zero-shot (unseen during training) datasets. Current state-of-the-art methods are trained on corresponding testing datasets, while other methods have never seen them during training. Our method achieves better performance than existing state-of-the-art methods on CamVid, ScanNet, and WildDash1, and is on par with existing methods on others. Compared with Mseg and HRNet trained on a single dataset, our methods is more robust in zero-shot testing. H-mean represents the Harmonic mean for all datasets. \label{Tab: Robustness cmp}}
\end{table*}

\begin{figure*}[!h]
\setlength{\belowcaptionskip}{-0.2cm}
\centering  %
\includegraphics[width=0.8\textwidth]{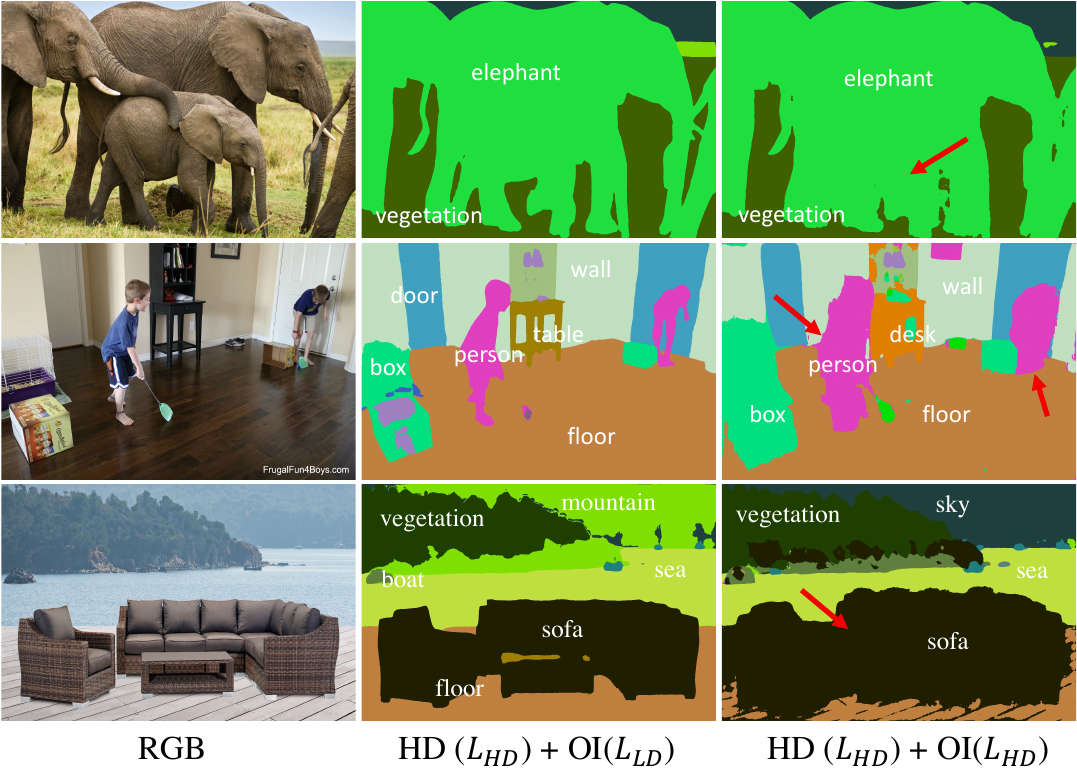}
\caption{Effectiveness of heterogeneous loss. We test on some google images. With the proposed heterogeneous loss, the segmented results are much more accurate. Red Arrows highlight some comparisons.  } 
\label{Fig: Effect of heterogeneous loss}
\end{figure*}

\noindent\textbf{Evaluation Metrics.} We employ the mean intersection over union (mIoU) to evaluate semantic segmentation and AP~\cite{lin2014microsoft} for the instance segmentation. Following~\cite{Ranftl2020}\cite{Wei2021CVPR}, we take absolute relative error (AbsRel) and percentage of pixels satisfying:
\[ \delta_{\tau} =  \max\left ( \frac{d_{pred}}{d_{gt}}, \frac{d_{gt}}{d_{pred}} \right )<\tau 
\]
for evaluation.  $d_{gt}$ and $d_{pred}$ are ground-truth and predicted depth respectively.

\noindent\textbf{Multi-scale Evaluation for Semantic Segmentation.} When evaluating the semantic segmentation performance, we employ multi-scale testing. We resize the testing images into multiple scales to feed into the model, i.e. with scales of 0.5 to 1.75 with 0.25 increments. Then we average the scores as the final prediction.

\noindent\textbf{Implementation Details.} We use two network architectures in our experiments, HRNet-W48~\cite{SunXLW19} and Segformer~\cite{xie2021segformer}. When training HRNet, we use SGD with momentum and polynomial learning rate decay, starting with a learning rate of 0.01. For the Segformer network, we use AdamW with polynomial learning rate decay and an initial learning rate 0.00006. To train a robust model on multiple datasets, following \cite{yin2021virtual}, we balance all datasets in a mini-batch to ensure each dataset accounts for an almost equal ratio. During training, images from all datasets are resized such that their shorter edge is resized to 1080. We randomly crop the image by $713\times713$ for HRNet and by $640\times640$ for Segformer. Other data augmentation methods are also applied, including random flip, color transformations, image blur, and image corruption. In the inference, we resize the image with the short edge to one of three resolutions (480$/$720$/$1080). When comparing with existing methods, multi-scale (ms) or single-scale (ss) testing is employed. %

\begin{table*}[!htb]
    \begin{minipage}{.65\textwidth}
      \centering
        \resizebox{\linewidth}{!}{%
\begin{tabular}{cccccccc|c}
\hline\hline
\multicolumn{1}{l|}{Methods}  & COCO   & ADE & Mapillary & IDD     & BDD     & Cityscapes & SUN & H-Mean\\ \hline
\multicolumn{1}{l|}{Mseg}  & 48.6   & 42.8   & 51.9      & 61.8    & 63.5    & 76.3       & 46.1  & 53.9  \\ 
\multicolumn{1}{l|}{Ours\_HRNet} & 53.3  & 48.4   & 55.3     & 65.0    & 69.0    & 79.2       & 50.5  &58.4 \\ 
\multicolumn{1}{l|}{Ours\_Segformer}  & \textbf{64.6}   & \textbf{55.1}   & \textbf{59.1} & \textbf{67.0}    & \textbf{70.5}    & \textbf{82.9}       & \textbf{50.5}   &\textbf{62.8} \\ \hline\hline
\end{tabular}}
\caption{Quantitative comparison of our method with Mseg on the validation set of training data. Our method achieves better performance than them over all datasets. \label{Tab: Cmp on val set.}}
    \end{minipage}%
    \hspace{0.2 cm}
    \begin{minipage}{.3\textwidth}
     \centering%
       \resizebox{0.5\linewidth}{!}{%
        \begin{tabular}{l|c}
\hline\hline
      & mIoU \\ \hline
Mseg~\cite{MSeg_2020_CVPR}  & 34.6 \\
Porzi~\cite{Porzi_2019_CVPR} & 37.1 \\ \hline
Ours  & \textbf{37.9} \\ \hline\hline
\end{tabular}}
\caption{Comparison with state-of-the-art methods on Wilddash2.\label{Tab: Cmp on WD2}}
    \end{minipage} 
\end{table*}

\subsection{Semantic Segmentation Evaluation}
To demonstrate the robustness and effectiveness of our methods, we conduct evaluation on 15  datasets.

\begin{table*}[]
\centering
\resizebox{0.99\linewidth}{!}{%
\begin{tabular}{ r llllllll}
\hline\hline
\multicolumn{1}{l|}{Methods}      & Arch.     & Training Data & CamVid & KITTI  & VOC      & Pascal Context & ScanNet & Wildash1  \\ \hline
\multicolumn{1}{l|}{Ours}         & HRNet-W48 & HD             & 82.8 & 63.8  & 71.6 &45.8 & 46.3 & 63.4    \\
\multicolumn{1}{l|}{Ours} & HRNet-W48 & HD+OB  & 83.4   & 66.0   & 74.6      &48.3   & 49.8   & 63.4   \\
\multicolumn{1}{l|}{Ours} & HRNet-W48 & HD+OI  & 83.4   & 66.9   & 74.9   & 48.6   & 52.0   & 64.0     \\ 
\multicolumn{1}{l|}{Ours}         & Segformer-B5 & HD   & 83.6   & 66.5   & 79.9      &53.0  & 53.0    & 68.2  \\
\multicolumn{1}{l|}{Ours}         & Segformer-B5 & HD+OI+OB  & \textbf{83.7}   & \textbf{68.9} & \textbf{81.1} & \textbf{54.2} & \textbf{55.3}& \textbf{69.7}   \\ \hline\hline
\end{tabular}}\\
\caption{Effectiveness of add weakly-annotated data and noisy data with our proposed heterogeneous losses. \label{Tab: effectiveness of heterogeneous loss.}}
\end{table*}

\noindent\textbf{Robustness Evaluation.}
To show the robustness of our method, we firstly compare with existing state-of-the-art methods on 6 zero-shot datasets (unseen to our method during training). Note that current methods have been trained on testing datasets. Results are shown in Table~\ref{Tab: Robustness cmp}. Our method can achieve state-of-the-art performance on CamViD, ScanNet, and WildDash1. Besides, we list the performance of HRNet~\cite{SunXLW19} trained on an individual training dataset. We can see that our method, trained on mixed datasets, shows much better robustness than them. Furthermore, Mseg~\cite{MSeg_2020_CVPR} laboriously merged 7 well-annotated datasets and craftily designed a unified taxonomy for labels mapping. It shows strong generalization over these zero-shot testing datasets. By contrast, our methods, using sentences to automatically establish the label's closeness, achieves much better performance than them over all zero-shot testing data. Furthermore, we show the performance comparison on the validation set of training data in Table~\ref{Tab: Cmp on val set.}. Our well-trained model outperforms Mseg over all datasets.

\noindent\textbf{Evaluation on Wilddash2.}
The Wilddash2 benchmark is intended for testing the robustness of models trained on other datasets, and does not provide a training set of its own. Our method achieves the state-of-the-art performance on it. Results are shown in Table~\ref{Tab: Cmp on WD2}.

\noindent\textbf{Effectiveness of Aggregating Noisy and Weak annotations with Heterogeneous Losses.}
We propose the heterogeneous losses to supervise the merged datasets. When aggregating OpenImages (`HD+OI') supervised with our $L_{\rm HD}+L_{\rm LD}$, Table~\ref{Tab: effectiveness of heterogeneous loss.} shows the performance can be improved consistently over all zero-shot datasets. Figure~\ref{Fig: Effect of heterogeneous loss} qualitatively compares with or without $L_{\rm LD}$ supervision for OpenImages (OI). We collect several web images for evaluation. We can observe that without $L_{\rm LD}$ loss (`HD ($L_{\rm HD}$) + OI ($L_{\rm HD}$)'), the predicted segments are much worse. Furthermore, when merging the Objects365 (OB) with the loss $L_{\rm WD}$, the performance is improved over all zero-shot datasets. %

\begin{table}[]
\centering
\resizebox{0.98\linewidth}{!}{%
\begin{tabular}{l|rrrrr|r}
\hline\hline
\multicolumn{1}{c|}{\multirow{2}{*}{Method}} & Lizard & Shark & Frog & Deer &duck & Mean \\\cline{2-7} 
\multicolumn{1}{c|}{}     & \multicolumn{6}{c}{mIoU (ss)}  \\ \hline
JoEm~\cite{baek2021exploiting}    &6.3   &20.3    &15.9  &5.1 &11.5 &11.8 \\
Mseg~\cite{MSeg_2020_CVPR}    &-   &-    &-  &- &- &17.3 \\
Ours (Word)    &13.1   & 11.2   &11.1  &4.2  &22.8 &12.5 \\
Ours (Sentence)    &23.6   & 51.6   &25.8  &17.8 &36.1 &31.0  \\ \hline\hline
\end{tabular}}
\caption{Comparison of zero-shot labels on the YoutubeVIS dataset. We compare with Mseg and a zero-shot semantic segmentation method, JoEm. As these labels do not exist in Mseg predetermined categories, we can only evaluate their animals mask accuracy on these labels. By contrast, JoEm and our method can segment these zero-shot labels. Our method, using the sentence description to create language embeddings, can achieve much better performance than others.
\label{Tab: zero-shot exp.}}
\end{table}

\noindent\textbf{Generalization on Unseen Labels.} As we employ language embeddings to represent semantic labels, the visual similarity between labels are established by text expressions.  In this experiment, we aim to observe if our resulting model can segment some zero-shot categories. Note that all these testing labels have never existed in our training data. We sampled around $2100$ images with 5 zero-shot labels from YoutubeVIS~\cite{Yang2019vis}.
Under this setting, we mainly compare with Mseg and JoEm~\cite{baek2021exploiting} because Mseg has strong generalization over cross domains and JoEm is the current state-of-the-art zero-shot semantic segmentation methods~\cite{baek2021exploiting}. We use their released weight for testing. JoEm is trained on the Pascal Context to segment 6 unseen classes. 
In contrast, as these labels are not in Mseg predetermined categories, we can only merge all their predicted animal labels to a single mask and evaluate the merged 'animal' mask accuracy. We cannot retrieve any category information from its results. By contrast, ideally, our method and JoEm can segment these zero-shot labels, since we can encode descriptions of labels in language embeddings. By comparing the cosine similarity between predicted embeddings and language embeddings, the semantic label can be retrieved. We compare two encoding methods: 1) encoding label words (`Ours (Word)') 2) encoding short descriptions
(`Ours (Sentence)').  Results are illustrated in Table~\ref{Tab: zero-shot exp.}. Our method is more robust and can achieve better performance than JoEm and Mseg. Furthermore, we can observe that using sentences can achieve much better performance.

\noindent\textbf{Finetuning on Small Datasets.}
In this experiment, we fine-tune our model on small datasets to show our well-trained weight can boost the performance significantly. We compare our well-trained weights, HRNet-W48 and Segformer-B5, with other weights on NYUv2~\cite{silberman2012indoor} and Pascal Context~\cite{mottaghi2014role}. We fine-tune the model for 100 epochs on NYU and 50 epochs on Pascal Context. Quantitative comparisons are shown in Table~\ref{Tab: Finetune on small datasets.}. Compared with ImageNet pre-trained weight `*(ImageNet)', our method surpasses them by a large margin, over 5\% higher. Similarly, our method is better than Mseg and ADE20K pretrained weight (`*(Mseg)', `*(ADE20K)'). Compared with state-of-the-art methods, our method can achieve much better performance.  

\noindent\textbf{Effectiveness of Distillation.} To employ large-scale weakly annotated data (Object365), we propose to use CLIP to distill our model. The experiment results are reported in Tab.~\ref{Tab: effectiveness of heterogeneous loss.}. `HRNet-W48 HD' is the baseline, which trains the model on large-scale high-quality semantic segmentation data. `HRNet-W48 HD+OB' merge the segmentation data with object detection data (Object365). Although we only roughly distill the knowledge on cropped bounding box regions, the performance can still achieve noticeable improvement. We hypothesize the gains come from the large-scale Object365 data and CLIP's  strong and robust semantic knowledge.

\noindent\textbf{Effectiveness of Merging Training Data with Language Embeddings.} How to merge training data labels is important for mixed data training. In this ablation, we compare two different approaches to solve this problem. The baseline method is to naively merge all data labels and employ a cross-entropy loss to supervise the training, see `Naive Merge' in Tab.~\ref{Tab: Cmp diff merge.}. We can observe that the performance on zero-shot datasets is much worse than that of Mseg, which proposes a unified taxonomy for manually mapping labels. In contrast, we propose to employ sentence embeddings to represent labels. As the sentence embeddings can represent the underlying semantic similarities between classes, which is more applicable to solving label conflicts. Our method can achieve better performance than others.

\newcolumntype{R}{>{\color{black}}r}
\begin{table}[]
\resizebox{0.97\linewidth}{!}{%
\begin{tabular}{R|RRRRRR}
\hline
             & VOC  & PC   & CamVid & WildDash 1 & KITTI & ScanNet \\ \hline \hline
Naive Merge  & 17.8 & 19.4 & 56.3   & 55.9       & 61.6  & 45.6    \\
Human Effort & 70.8 & 45.2 & 82.4   & 64.2       & 62.4  & 48.4    \\
Ours & 74.9 & 48.6 & 83.4   & 64         & 66.9  & 52      \\ \hline\hline
\end{tabular}}
\caption{Comparison of data merging methods for training. All the models are trained with the same network structure of HRNet. When mixing 7 high-quality data for training, we compare with the `Naive Merge' and `Human Effort'. `Naive Merge' does not solve the label conflicts when mixing multiple datasets, while `Human Effort' employs a unified taxonomy for manual mapping as stated in~\cite{MSeg_2020_CVPR}. In contrast, we employ sentence embeddings to represent labels, which can achieve better performance on mixed data training with less human effort. Note that all testing data are unseen during training. }
\label{Tab: Cmp diff merge.}
\end{table}

\begin{figure*}[!h]
\setlength{\belowcaptionskip}{-0.2cm}
\centering  %
\includegraphics[width=0.8\textwidth]{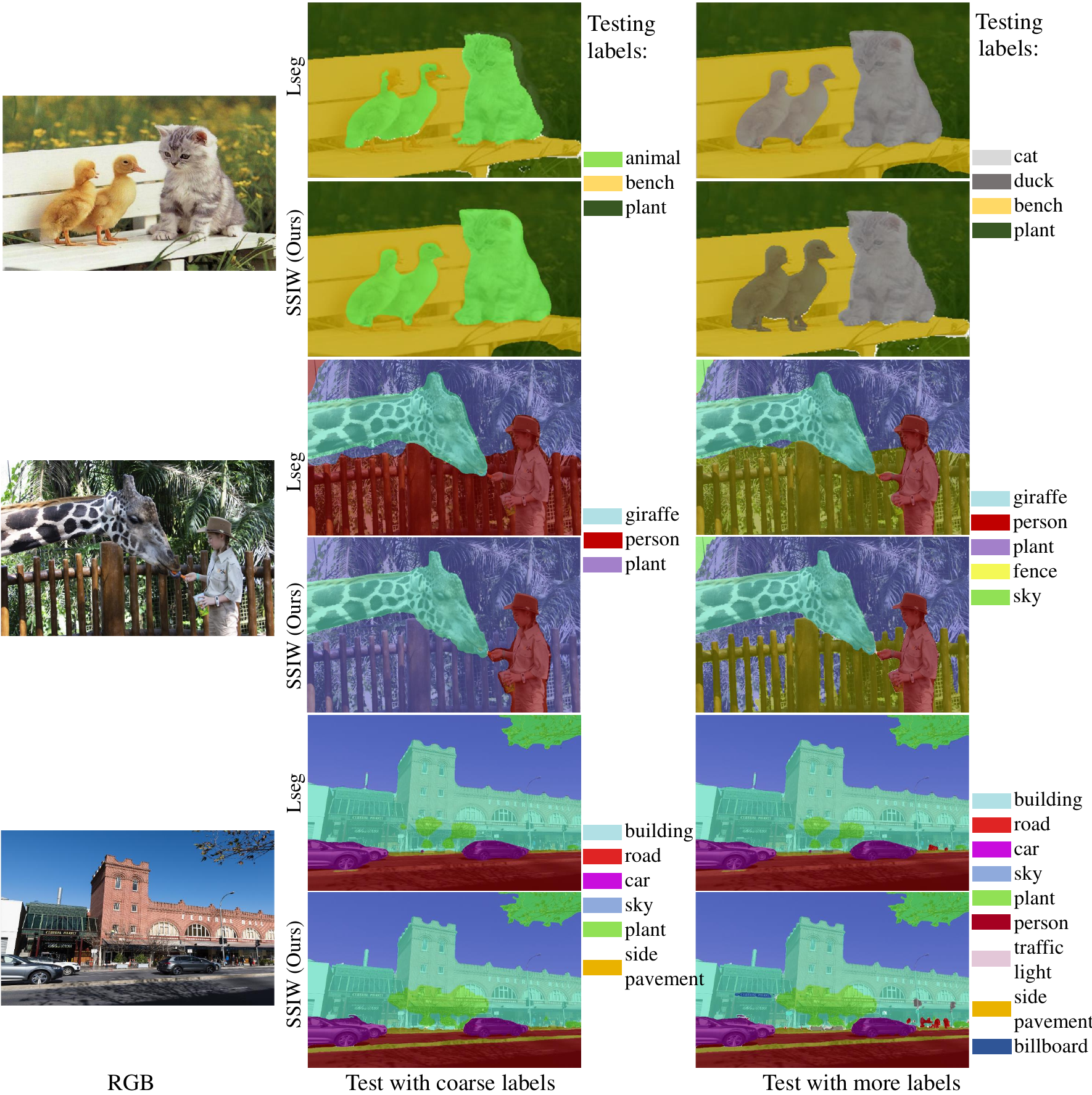}
\caption{Comparison with Lseg~\cite{li2022languagedriven} on web images. We collect several online photos and test them with different label sets. We test on two different label sets, i.e. one has coarse labels, while the other one has more refined labels. Our method can segment better on the refined label set.}
\label{Fig: cmp with lseg}
\end{figure*}

\noindent\textbf{Comparison with Lseg}
The co-existing approach, Lseg~\cite{li2022languagedriven}, proposes a solution to perform zero-shot semantic segmentation by incorporating CLIP feature in their pipeline. In contrast, our method employs CLIP to create soft embeddings for sentences, allowing us to address the mixed-data taxonomy and enhance the model's ability to generalize to zero-shot domains and labels by scaling up the training data. A quantitative comparison is presented in Fig.~\ref{Fig: cmp with lseg}, where the model is evaluated on different label sets with varying levels of granularity. While both methods demonstrate comparable performance on coarse labels, our approach produces superior segmentation results on refined labels.

\begin{table*}[]
\label{Tab: Finetune on small datasets.}
\centering
\begin{tabular}{llll}
\hline\hline
\multicolumn{2}{l|}{NYUDV2}      & \multicolumn{2}{l}{Pascal Context} \\ \hline
\multicolumn{4}{c}{mIoU (ss)}                                                                                                      \\ \hline
MTI-Net~\cite{vandenhende2020mti}          & \multicolumn{1}{l|}{49.0}   & OCR~\cite{yuan2020object}                    & 59.6      \\
ICM~\cite{shi2019scene}            & \multicolumn{1}{l|}{50.7} & CAA~\cite{huang2021channelized}                    & 60.5      \\
ShapeConv~\cite{cao2021shapeconv} & \multicolumn{1}{l|}{51.3} & DPT~\cite{ranftl2021vision}                    & 60.5      \\ \hline
HRNet (ImageNet)   & \multicolumn{1}{l|}{44.5} & HRNet~\cite{SunXLW19}       & 50.3      \\
HRNet (Mseg)        & \multicolumn{1}{l|}{53.4} & HRNet (Mseg)            & 52.6      \\
HRNet (Ours)    & \multicolumn{1}{l|}{54.1} & HRNet (Ours)            & 52.8      \\ \hline
Segformer (ImageNet)            & \multicolumn{1}{l|}{50.0} & Segformer (ImageNet)              & 59.4      \\
Segformer (ADE20K)         & \multicolumn{1}{l|}{59.1} & Segformer (ADE20K)      & 60.8      \\
Segformer (Ours)       & \multicolumn{1}{l|}{\textbf{60.0}} & Segformer (Ours)        & \textbf{65.0}      \\ \hline\hline
\end{tabular}
\caption{Fine tuning on small datasets. Our well-trained weight can provide strong performance on NYU and Pascal Context, which surpasses the state-of-the-art methods by a large margin. }
\end{table*}

\begin{table*}[t]
\centering
\setlength{\tabcolsep}{2pt}
\resizebox{0.98\linewidth}{!}{%
\begin{tabular}{l|l|ll|ll|ll|ll|ll}
\hline\hline
\multirow{2}{*}{Method} & \multirow{2}{*}{Backbone} & \multicolumn{2}{c|}{NYU} & \multicolumn{2}{c|}{KITTI} & \multicolumn{2}{c|}{DIODE} & \multicolumn{2}{c|}{ScanNet}  & \multicolumn{2}{c}{Sintel} \\
&    & AbsRel$\downarrow$     & $\delta_{1}\uparrow$     & AbsRel$\downarrow$      & $\delta_{1}\uparrow$      & AbsRel$\downarrow$      & $\delta_{1}\uparrow$      &AbsRel$\downarrow$      & $\delta_{1}\uparrow$       &AbsRel$\downarrow$     & $\delta_{1}\uparrow$   \\ \hline
OASIS~\cite{chen2020oasis}  &ResNet50  &$21.9$ &$66.8$ &$31.7$ & $43.7$ &  $48.4$ &$53.4$ &$19.8$ &$69.7$ &$60.2$ &$42.9$ \\ 
MegaDepth~\cite{li2018megadepth}& Hourglass    &$19.4$& $71.4$ &$20.1$ &$66.3$ &$39.1$ &$61.5$ &$19.0$ &$71
.2$ &$39.8$ &$52.7$ \\
Xian \etal \cite{xian2020structure} &ResNet50  &$16.6$ &$77.2$ & $27.0$  & $52.9$ &$42.5$ &$61.8$ &$17.4$ &$75.9$ &$52.6$ &$50.9$  \\
DiverseDepth~\cite{yin2020diversedepth}\cite{yin2021virtual}&ResNeXt50  &$11.7$ &$87.5$ &$19.0$ &$70.4$ &$37.6$ &$63.1$ &$10.8$ &$88.2$  &$38.6$ &$58.7$  \\
MiDaS~\cite{Ranftl2020}&ResNeXt101 &$11.1$ &$88.5$ &$23.6$ &$63.0$ &$33.2$ &$71.5$ &$11.1$ &$88.6$ &$40.5$ &$60.6$\\
LeReS~\cite{Wei2021CVPR} &ResNet50  &$9.1$  &$91.4$  &$14.3$ &$80.0$ &$28.7$ &$75.1$ &$9.6$ &$90.8$ &$34.4$ &$62.4$   \\     
\hline
Ours $+$ LeReS &ResNet50  &$\textbf{8.6}$  &$\textbf{92.3}$  &$\textbf{14.0}$ &$\textbf{80.6}$ &$\textbf{27.4}$ &$\textbf{75.8}$ &$\textbf{8.0}$ &$\textbf{93.4}$ &$\textbf{29.2}$ &$\textbf{62.4}$
\\ \hline\hline
\end{tabular}}
\caption{Quantitative comparison of our depth prediction with state-of-the-art methods on five zero-shot (unseen during training) datasets. Our method input with pseudo semantic labels achieves much better performance than existing state-of-the-art methods across all test datasets. \label{Tab: Depth cmp}}
\end{table*}

\begin{table*}[h!]
\centering
\resizebox{0.9\linewidth}{!}{%
\begin{tabular}{ l | l |c|c|cc|ccc}
\hline\hline
Method & Backbone & Schedule & AP (\%) & AP$_{50}$ & AP$_{75}$ & AP$_{S}$ & AP$_{M}$ & AP$_{L}$ \\ \hline
Mask R-CNN~\cite{he2017mask} & R-50-FPN  & $3\times$ & 37.5 & 59.3 & 40.2 & \textbf{21.1} & 39.6 & 48.3 \\
TensorMask~\cite{chen2019tensormask} & R-50-FPN  & $6\times$ & 35.4 & 57.2 & 37.3 & 16.3 & 36.8 & 49.3 \\
BlendMask~\cite{chen2020blendmask} & R-50-FPN  & $3\times$ & 37.0 & 58.9 & 39.7 & 17.3 & 39.4 & 52.5\\ \hline
CondInst (10\%COCO$^*$) & R-50-FPN  & $1\times$ & 21.7 & 35.2 & 22.0 & 8.9 & 22.4 & 31.1 \\
\textbf{Ours (CondInst), 10\%COCO$^*$} & R-50-FPN  & $1\times$ & 33.2 & 52.1 & 35.2 & 15.0 & 36.4 & 48.0 \\ \hline
CondInst (25\%COCO$^*$) & R-50-FPN & $1\times$ & 31.4 & 50.6 & 33.1 & 14.5 & 34.2 & 45.9 \\
\textbf{Ours (CondInst), 25\%COCO$^*$} & R-50-FPN  & $1\times$ & 36.4 & 56.9 & 38.6 & 17.7 & 39.4 & 53.0 \\ \hline
CondInst~\cite{tian2020conditional} & R-50-FPN  & $1\times$ & 35.9 & 57.0 & 38.2 & 19.0 & 40.3 & 48.7 \\
\textbf{Ours (CondInst), 50\%COCO$^*$} & R-50-FPN  & $1\times$ & 38.1 & 59.0 & 40.8 & 18.7 & 41.5 & 55.2 \\
\textbf{Ours (CondInst), COCO} & R-50-FPN & $1\times$ & \textbf{39.5} & \textbf{60.7} & \textbf{42.5} & 19.5 & \textbf{42.9} & \textbf{57.5} \\
\hline \hline
Mask R-CNN & R-101-FPN & $6\times$ & 38.3 & 61.2 & 40.8 & 18.2 & 40.6 & 54.1 \\
PolarMask~\cite{xie2020polarmask} & R-101-FPN  & $2\times$ & 32.1 & 53.7 & 33.1 & 14.7 & 33.8 & 45.3 \\
TensorMask & R-101-FPN  & $6\times$ & 37.1 & 59.3 & 39.4 & 17.4 & 39.1 & 51.6 \\
BlendMask & R-101-FPN  & $3\times$ & 39.6 & 61.6 & 42.6 & \textbf{22.4} & 42.2 & 51.4 \\
SOLOv2~\cite{wang2020solov2} & R-101-FPN  & $6\times$ & 39.7 & 60.7 & 42.9 & 17.3 & 42.9& 57.4\\
CondInst & R-101-FPN  & $3\times$ &40.0 & 62.0 & 42.9 & 21.4 & 42.6 & 53.0 \\
CondInst & R-101-BiFPN  & $3\times$ & 40.5 & 62.4 & 43.4 & 21.8 & 43.3 & 53.3 \\\hline
\textbf{Ours (CondInst)} & R-101-BiFPN  & $3\times$ &  \textbf{41.4} &  \textbf{63.0} &  \textbf{44.7} & 20.0 &  \textbf{45.5} &  \textbf{59.5} 
 \\\hline\hline
\end{tabular}}\\
\caption{Quantitative comparison of our instance segmentation with state-of-the-art methods on COCO. With our created pseudo labels on Objects365, the performance is improved significantly. $^*$ Used 10\% $\sim$ 50\% of COCO data for training.}
\label{Tab: Inst cmp}
\end{table*}

\subsection{Boosting Downstream Applications}

\begin{figure}[]
\setlength{\belowcaptionskip}{-0.4cm}
\centering  %
\includegraphics[width=0.48\textwidth]{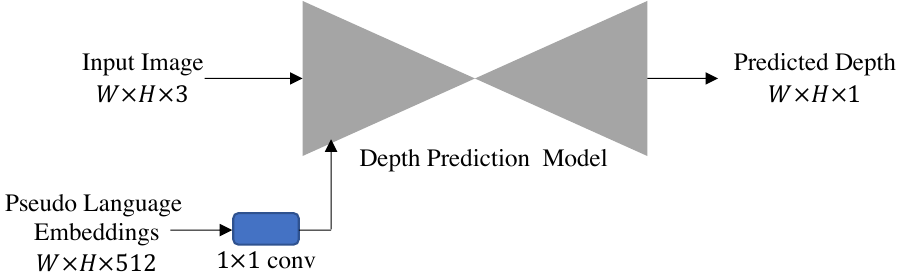}
\caption{\textbf{Depth Prediction Framework}. The language embeddings are input to the depth prediction network.} 
\label{Fig: depth prediction framework}
\end{figure}

\noindent\textbf{Boosting  Monocular Depth Estimation.}
We create pseudo semantic labels on multiple zero-shot depth datasets to boost monocular depth estimation. Following LeReS~\cite{Wei2021CVPR}, we train monocular depth estimation model on Taskonomy~\cite{zamir2018taskonomy}, DIML~\cite{kim2018deep}, Holopix~\cite{hua2020holopix50k}, and HRWSI~\cite{xian2020structure} and evaluate it on 5 zero-shot datasets. On these 9 unseen datasets, we create pseudo semantic masks and transfer them to pixel-wise language embeddings, whose dimensions are 512. During training, we employ the ResNet50
backbone. The language embeddings are feed to a $1 \times 1$ convolution to reduce the dimension from $512$ to $64$, which is added to the depth prediction network. Then we add the transferred embeddings to the image feature after the first $7\times 7$ convolution. The framework is shown in Fig.~\ref{Fig: depth prediction framework}.
Results are shown in  Table~\ref{Tab: Depth cmp}. Comparing with the baseline method `LeReS', adding our created embeddings can consistently improve the performance over all datasets.

\noindent\textbf{Boosting Instance Segmentation.}
Instance segmentation is another important fundamental problem. To demonstrate that our method can boost its performance, we create pseudo instance masks on sampled Objects365. The created pseudo labels have 7000K instance masks and 73 categories. We firstly train CondInst~\cite{tian2020conditional} on pseudo labels and then finetune it on COCO~\cite{lin2014microsoft}. Comparisons are illustrated in Table~\ref{Tab: Inst cmp}. With the ResNet-50 backbone, our method (`Ours(CondInst), 25\%COCO'), using only 25\% COCO data, can achieve comparable performance with the baseline method `CondInst', and can surpass other state-of-the-art methods with 50\% COCO data. When finetuned on the whole COCO, our method is (`Ours(CondInst), COCO') around 4\% AP higher than the baseline (`CondInst'), and is even better than many methods with ResNet-101 backbone. We can observe that the main improvement comes from the medium ($AP_{M}$) and large ($AP_{L}$) objects.

\section{Discussion}
\noindent\textbf{Limitations.} We also observe a few limitations of our method. First, the performance of the model may be limited by the representation of the language model. Zero-shot categories with very similar semantic expressions may be confused.
Second, the model may fail to generalize to classes that are too far from the training language space. 
However, we %
believe 
that if merging data with more categories in our method, we can achieve more distinguishable features and alleviate this problem. The improvement of the language model may also benefit the proposed method. 

 \noindent\textbf
{Conclusion.} In this work, we have 
proposed an approach to semantic segmentation that can achieve promising performance over multiple zero-shot cross-domain datasets. By collecting short label descriptions from Wikipedia and encoding them in vector-valued embeddings to replace labels, we can easily merge multiple datasets together to retrieve a strong and robust segmentation model. A heterogeneous loss is proposed to leverage noisy datasets and weakly annotated datasets. Extensive experiments demonstrate that we can achieve better or comparable performance with current state-of-the-art methods on 7 cross-domain datasets. Furthermore, our resulting model demonstrates the ability to segment some zero-shot labels. With our robust and strong model, the downstream applications can also be boosted significantly.  

\noindent\textbf{Acknowledgements} Part of this work was done when Wei Yin was an intern at Amazon.

\bibliographystyle{IEEEtran}
\bibliography{reference}

\end{document}